\pdfoutput=1

\documentclass[11pt]{article}

\usepackage{acl2023}

\usepackage{times}
\usepackage{latexsym}

\usepackage[T1]{fontenc}

\usepackage[utf8]{inputenc}

\usepackage{microtype}

\usepackage{inconsolata}

\usepackage{tcolorbox}
\usepackage{soul}
\usepackage{float}
\usepackage{multirow}
\usepackage{multicol}
\usepackage{rotating}
\usepackage{graphicx}
\usepackage{subcaption} 

\title{Exploring Zero and Few-shot Techniques for Intent Classification}

\author{Soham Parikh \\
  ServiceNow Inc \\
  \texttt{soham.parikh@servicenow.com} \\
  \And
  Quaizar Vohra \\
  ServiceNow Inc \\
  \texttt{quaizar.vohra@servicenow.com} \\
  \AND
  Prashil Tumbade \\
  ServiceNow Inc \\
  \texttt{prashil.tumbade@servicenow.com} \\
  \And
  Mitul Tiwari \\
  ServiceNow Inc \\
  \texttt{mitul.tiwari@servicenow.com} \\}

\begin{document}
\maketitle

\begin{abstract}

Conversational NLU providers often need to scale to thousands of intent-classification models where new customers often face the cold-start problem. Scaling to so many customers puts a constraint on storage space as well. In this paper, we explore four different zero and few-shot intent classification approaches with this low-resource constraint: 1) domain adaptation, 2) data augmentation, 3) zero-shot intent classification using descriptions large language models (LLMs), and 4) parameter-efficient fine-tuning of instruction-finetuned language models. Our results show that all these approaches are effective to different degrees in low-resource settings. Parameter-efficient fine-tuning using T-few recipe \cite{liu2022fewshot} on Flan-T5 \cite{DBLP:journals/corr/abs-2210-11416} yields the best performance even with just one sample per intent. We also show that the zero-shot method of prompting LLMs using intent descriptions is also very competitive. 

\end{abstract}

\section{Introduction}

Intent classification is the primary natural language understandin task for a virtual agent or a chatbot. 
Providing intent-utterances for training intent classification models is a laborious process.
In this paper, we address this problem by exploring zero and few-shot intent identification using Large Language Models (LLMs) as well as instruction finetuned models. Zero-shot and few-shot intent prediction completely remove or substantially reduce the work to provide intent-utterances, respectively. We demonstrate that the following four approaches work well in practice for zero/few-shot intent classification. 
\begin{itemize}
\itemsep0em 
    \item \textbf{Domain adaptation}: We use a sentence encoder that is pre-trained with our domain knowledge and show that it performs well in a few-shot setting compared to off-the-shelf sentence encoders.

    \item \textbf{Data Augmentation} by supplementing human-curated training data with LLM-generated data to improve training data.
    
    \item \textbf{Zero-shot intent classification}: high capacity LLMs can be prompted creatively with intent descriptions to do zero-shot classification. 

    \item \textbf{Parameter-efficient fine-tuning (PEFT)}: finetuning a small number of parameters added to instruction finetuned LMs using only a few examples
\end{itemize}
Here is the outline of the rest of the paper. In Section~\ref{sec:related_work} we describe the related work. In Section~\ref{sec:datasets} we detail the datasets used. In Section~\ref{sec:methodology} we describe the four approaches covered in this work for zero/few-shot intent classification. Finally, we conclude with observations in Sections~\ref{sec:observations} and \ref{sec:conclusion}.

\begin{table*}[!h]
\centering
\small
\begin{tabular}{c|c|c|c|c}
\hline
    \textbf{Dataset} & \textbf{Intents} & \textbf{Train Size} & \textbf{Test Size} & \textbf{OOS Samples in Test}\\
    \hline
     MASSIVE & 60 & 11514 & 2974 & No \\
     OOTB-dataset* & 27 & 1363 & 3099 & No \\
     Benchmark01* & 9 & 270 & 300 & Yes \\
     Benchmark02* & 13 & 390 & 420 & Yes \\
     Benchmark03* & 31 & 930 & 960 & Yes
\end{tabular}
\caption{Statistics for intent classification datasets used in this paper. Datasets marked with an asterisk (*) are private, internal benchmarking datasets. Train and Test Sizes correspond to the number of utterances in the respective spits. OOS samples in test set indicates whether there are any out-of-scope samples in the test set.}
\label{tab: datasets}
\end{table*}
\section{Related Work}
\label{sec:related_work}

Recent papers have used domain adaptation \cite{yu-etal-2021-shot} and contrastive learning 
\cite{zhang-etal-2021-shot} successfully for few-shot intent classification. Yu et al.~\cite{yu-etal-2021-shot} use embeddings from a BERT model \cite{devlin2018pretraining} pretrained on domain data to search for utterances belonging to new intents in the domain. \cite{zhang-etal-2021-shot} finetune a BERT model on few-shot data using contrastive learning which learns to discriminate between semantically similar sentences. Our work on domain adaptation differs from these mainly due to our setting which involves serving thousands of customers. For legal reasons, we cannot co-mingle data from these customers to pre-train a single model. Instead, we pre-train a sentence encoder based on an intent taxonomy and out-of-the-box intents, which consist of human generated synthetic data. In this setting, we can only train very lightweight models for each customer, e.g. a dense layer on top of a pre-trained sentence encoder.

Data Augmentation is another widely used technique to solve the problem of data scarcity. Recent work on data augmentation has focused on using multiple methods to improve model performance \cite{DBLP:journals/corr/abs-2206-05790}. \cite{sahu-etal-2022-data} use a prompt-based approach to generate labeled training data for intent classification using LLMs like GPT-3 \cite{NEURIPS2020_1457c0d6}. The quality of generated training data using LLMs is highly dependent on the prompts. In this work, we show various prompt-based approaches that generate diverse data for training and boost the performance of intent classifiers.

As the usage of conversational agents grows, it is important for them to generalize to new intents. Recent work has focused on performing zero-shot intent detection on unseen intents and domains. \cite{7178987, yazdani-henderson-2015-model} use additional knowledge from ontologies or attributes whereas \cite{liu-etal-2019-reconstructing} make modifications to capsule networks to generalize to unseen domains. \cite{DBLP:journals/corr/abs-1912-09297} use embeddings of intent descriptions to perform zero-shot detection of new intents and services. While these methods are effective, they all require training on an initial set of intents. Large Language Models (LLMs) like GPT-3 \cite{NEURIPS2020_1457c0d6} and more recently instruction finetuned models like \cite{DBLP:journals/corr/abs-2210-11416} have shown good zero-shot performance on newly seen tasks without any prior training data on those tasks. In this work, we show that these models are also effective for zero-shot intent classification using just intent descriptions. 
\section{Datasets}
\label{sec:datasets}
We use public and private intent classification datasets to benchmark different approaches. For evaluation on public dataset, we use the English train and test sets from MASSIVE for intent classification. MASSIVE contains utterances directed at a physical device spanning 60 intents and 18 domains. For more details on the MASSIVE dataset \cite{DBLP:journals/corr/abs-2204-08582}, we encourage readers to refer to their paper. 
We also use private benchmarking datasets internal to our company. These datasets contain various intents and utterances in the enterprise setting spanning 3 different domains: IT Service Management (ITSM), HR and Customer Service Management (CSM). The utterances are inspired by interactions between humans and chatbots and are typically queries from goal-oriented conversations where the user needs to resolve an issue. Additionally, some of these datasets also contain out-of-scope (OOS) utterances in their test set i.e., utterances that do not belong to any intent, in order to benchmark irrelevance detection of intent classification models. Table \ref{tab: datasets} shows statistics for different datasets used in our benchmarking. 

\section{Methodology}
\label{sec:methodology}
In this section, we describe the various methods we test for zero and few-shot learning. 

\begin{table*}[!t]
\centering
\footnotesize
\begin{tabular}{|l|c|c|c|c|c|}
\hline
\textbf{Few-shot K} & \textbf{model} & \textbf{Massive} & \textbf{Benchmark01} & \textbf{Benchmark02} & \textbf{Benchmark03} \\
\hline
\multirow{4}{*}{3} & LaBSE & 46 (1.7) & 59 (2.9) & 52 (2.7) & 58 (3.1)\\
 & MUSE3 & 53 (2.8) & 64 (3.8) & 62 (2.7) & 64 (1.3)\\
 & GTR-3b & \textbf{59 (1.4)} & 76 (1.4) &	\textbf{70 (3.3)} & \textbf{78 (2.2)}\\
 & ELMSE  & 57 (2.3) & \textbf{77 (2.4)} & 63 (4.6)	& 74 (1.7)\\
\hline
\multirow{4}{*}{5} & LaBSE & 58 (1.7) &	65 (3.3) &	59 (1.7) &	67 (1.8)\\
 & MUSE3 &  61 (0.9) & 70 (2.2) & 66 (1.4) &	70 (1.7)\\
 & GTR-3b & \textbf{66 (1.2)} & 78 (1.0) & \textbf{73 (1.7)} &	\textbf{84 (1.0)} \\
 & ELMSE & 63 (1.1) & \textbf{80 (1.7)} &	67 (2.6) &	79 (1.2) \\
\hline
\end{tabular}
\caption{Results for domain adaptation on 3 internal datasets along with MASSIVE comparing LaBSE, MUSE, ELMSE, and GTR-3B models. The metric reported here is in-scope accuracy averaged over 5 different selections of few shot data. Numbers inside parenthesis indicate standard deviation across the 5 selections}
\label{tab:domain_adaptation_5shot}
\end{table*}
\subsection{Domain Adaptation}
\label{subsec:domain_adapt}
Domain and task-specific pre-training of language model \cite{gururangan-etal-2020-dont} has shown to significantly improve classification accuracy in both low and high resource settings. Techniques like contrastive learning \cite{gao-etal-2021-simcse} \cite{feng-etal-2022-labse} are effective for improving the quality of sentence encoders, specifically in low-resource settings. Inspired by these ideas, we use a sentence encoder trained on our domain-specific data along with public datasets. Starting with the LaBSE checkpoint ~\cite{feng-etal-2022-labse} we train it further by converting intent classification, paraphrasing, etc, as sentence similarity tasks. We will refer to this model as ELMSE (enterprise language model based sentence encoder). 

For training intent-classification models, we freeze ELMSE weights and use its sentence embeddings as features for a trainable non-linear dense layer for classification. We compare ELMSE against other publicly available sentence encoders, namely LaBSE, Multilingual Universal Sentence Encoder (MUSE) \cite{DBLP:conf/acl/YangCAGLCAYTSSK20} and GTR-3B. ELMSE is comparable in size to LaBSE and MUSE while almost 30 times smaller than GTR-3b.
We simulate few-shot setting by randomly selecting K utterances per intent from full datasets. We use K=3,5,8,10,15,20 as well as the full dataset. We report results on 4 datasets from Table \ref{tab: datasets}. Since OOTB-dataset was used for pretraining ELMSE, we exclude it from few-shot evaluation. 
\begin{figure}[!t]
\centering
\includegraphics[width=0.5\textwidth]{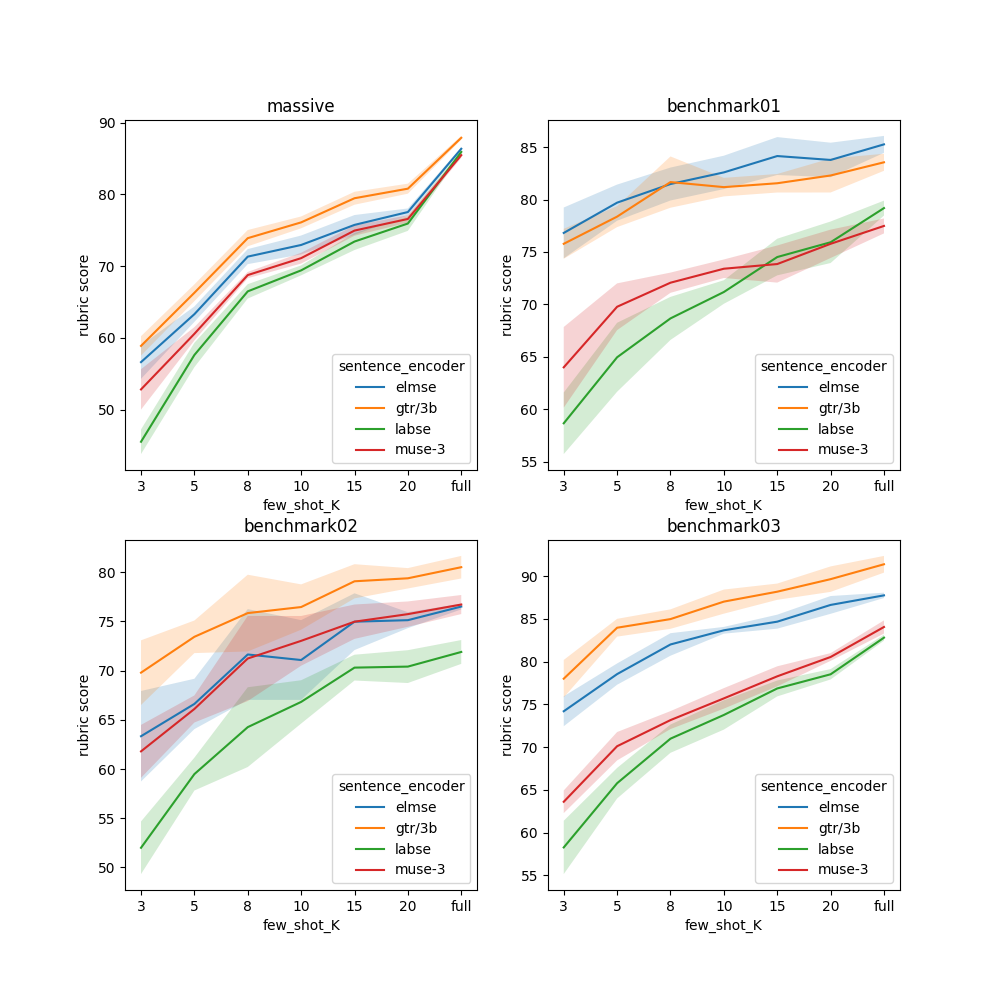}
\caption{Comparison of ELMSE which is domain adapted with sentence encoders which are not domain adapted}
\label{fig:domain_adaption}
\end{figure}

\subsubsection{Results for Domain Adaptation}
Table~\ref{tab:domain_adaptation_5shot} reports in-scope accuracy and standard deviation averaged of 5 random seeds for 3-shot and 5-shot classification. The results demonstrate that domain adaptation is a very effective approach with improvements of greater than 5 percent in most cases when compared with models of similar size. These results carry
over as we increase the number of few-shot utterances to more than 5 as shown in Figure~\ref{fig:domain_adaption}. The plots also show that the gap between ELMSE and LaBSE is much larger in a few-shot setting and reduces as K increases. Moreover, ELMSE is only 2-3\% worse than GTR-3b which is 30 times larger model. 

\begin{table*}[!t]
\centering
\scriptsize
\begin{tabular}{|c|c|c|}
\hline
Approach &
  Prompt &
  Generated Text \\ \hline
\begin{tabular}[c]{@{}l@{}}GPT3 Paraphrase  Aug.\end{tabular} &
  \begin{tabular}[c]{@{}l@{}}\textbf{Task}: Create diverse utterances\\ by paraphrasing the following utterances:\\ \textit{schedule alarm to wake me up after 3 hours}\\ \textit{alarm for ten am}\\ \textit{wake me up on friday at five in the morning i need to catch the train}\\ \textit{alarm me at eight am}\\ \textit{please set alarm for today}\\\textbf{Create 20 utterances:}\end{tabular} &
  \begin{tabular}[c]{@{}l@{}}Set an alarm for 10 o'clock.\\ \\ Wake me up on Friday at 5am\\ so I can make the train.\\ \vdots\\ Set a timer to wake me up \\ in three hours\end{tabular} \\ \hline
\begin{tabular}[c]{@{}l@{}}GPT3 Aug. Using Intent Descr.\end{tabular} &
  \begin{tabular}[c]{@{}l@{}}A virtual assistant serves multiple intents.\\ Below are the description of the intents:\\ \textbf{alarm\_set}: user wants to set an alarm\\ \textbf{iot\_cleaning}: user wants to do some cleaning \\ \vdots \\ \textbf{play\_podcasts}: user wants to play a podcast or rewind/repeat a particular \\ episode in a podcast \\ Generate \textbf{20 utterances for alarm\_set} intent:\end{tabular} &
  \begin{tabular}[c]{@{}l@{}}Can you set an alarm for next week?\\ I need to set an alarm for a specific time \\ I want to set an alarm for a certain day\\ \vdots\\ I'd like to set an alarm \\for a certain hour\end{tabular} \\ \hline
\end{tabular}
\caption{Example prompts used in generating text for the corresponding approaches }
\label{tab:prompts_table}
\end{table*}

\begin{table*}[!t]
\centering
\footnotesize
\begin{tabular}{l|c|c|c|c}
\hline
\multirow{1}{*}{\textbf{Approach}} & \multirow{1}{*}{\textbf{MASSIVE}} & \textbf{Benchmark01} & \textbf{Benchmark02} & \textbf{Benchmark03} \\
\cline{1-5}
\multirow{1}{*}{ELMSE Baseline} & \multirow{1}{*}{63 (1.1)} & 80 (1.7) & 67 (2.6) & 79 (1.2) \\

\multirow{1}{*}{GPT-3 w/ Paraphrase Aug.} & \multirow{1}{*}{\textbf{63 (0.5)}} & \textbf{84 (0.4)} & 71 (0.3) & \textbf{81 (0.5)} \\

\multirow{1}{*}{GPT-3 w/ Intent Descriptions} & \multirow{1}{*}{51 (0.5)} & 76 (0.4) & 69 (0.5) & 76 (0.2) \\


\multirow{1}{*}{Parrot T5} & \multirow{1}{*}{58 (0.4)} & 81 (0.2) & \textbf{73 (0.4)} & 81 (0.4) \\

\multirow{1}{*}{Seed Set + GPT-3 w/ Intent Descriptions} & \multirow{1}{*}{63 (0.8)} & 84 (0.4) & 71 (0.3) & 78 (0.9) \\

\multirow{1}{*}{Seed Set + Parrot T5} & \multirow{1}{*}{63 (0.6)} & 79 (0.4) & 68 (2.2) & 76 (0.6) \\

\cline{1-5}

\end{tabular}
\caption{Results for Data Augmentation on 3 internal datasets along with MASSIVE comparing the performance on multiple prompt-based approaches. We report the average in-scope accuracy and standard deviation averaged over 3 different random seeds}
\label{tab:data_augmentation}
\end{table*}

\subsection{Data Augmentation}
We use data augmentation to generate labeled data for training starting with a seed set of 5 utterances per intent. In this section, we exploer different ways of prompting GPT-3 and T5 \cite{DBLP:journals/jmlr/RaffelSRLNMZLL20}. For evaluating the generated utterances, we use them for training the same type of lightweight classifier as described in \ref{subsec:domain_adapt} using ELMSE as the sentence encoder. 
This section describes different prompt-based approaches for data generation.
\paragraph{GPT-3 + Paraphrase} Following \cite{sahu-etal-2022-data}, we ask GPT-3 to generate 20 paraphrases of utterances from the same intent taken from the seed set. To encourage diverse generations, we set high temperature and top\_p values.
\paragraph{GPT-3 + Intent Descriptions} We describe intents in the prompt and ask GPT-3 to generate 20 utterances for a particular intent. We find that describing all intents prevents hallucinations in the generations.
\paragraph{Parrot T5 Paraphrasing} We use the Parrot Paraphrase approach based on T5 \cite{prithivida2021parrot} to generate 20 diverse paraphrased utterances given seed set.
Table \ref{tab:prompts_table} shows a few generations from our prompt-based approaches.

\subsubsection{Experimental Setup and Results}
To evaluate the quality of generated utterances, we use them to train intent classifiers. We evaluate the performance of augmented dataset from each approach as mentioned in Section 4.2.1 by training ELMSE classifier model for intent classification task. We evaluate on 4 datasets and compared against ELMSE few-shot baseline where K is set to 5. We report the in-scope accuracy and standard deviation averaged over 3 different random seeds. Table \ref{tab:data_augmentation} shows the result for all approaches using the data augmentation. Unless mentioned explicitly, we do not add the seed set to the training mix. 

We find that using paraphrases from GPT-3 and Parrot T5 Paraphraser give better results compared to ELMSE Baseline even without the seed set. GPT-3 Augmentations using Intent Descriptions does not perform well but when combined with ELMSE Baseline seed set gives better results. Moreover, given a good quality seed-set, we see that data augmentation using LLMs can boost the performance of intent classification in few-shot setting.

\begin{table*}[!t]
\centering
\footnotesize
\begin{tabular}{l|c|c|c|c}
\hline
\textbf{Dataset} & \textbf{LLM Intents} & \textbf{Model} & \textbf{In-Scope Accuracy} & \textbf{Out-of-scope Recall} \\
\hline
\multirow{6}{*}{MASSIVE (60 intents)} & \multirow{2}{*}{5} & Flan-T5-XXL & 68.6 & - \\
& & GPT-3 & \textbf{69.2} & - \\
\cline{2-5}
& \multirow{2}{*}{60} & Flan-T5-XXL & 73.3 & - \\
& & GPT-3 & \textbf{73.9} & - \\
\cline{1-5}
\multirow{6}{*}{OOTB-dataset (27 intents)} & \multirow{2}{*}{5} & Flan-T5-XXL & \textbf{83.7} & - \\
& & GPT-3 & 83.4 & - \\
\cline{2-5}
& \multirow{2}{*}{27} & Flan-T5-XXL & \textbf{86.3} & - \\
& & GPT-3 & 84.9 & - \\
\cline{1-5}
\multirow{6}{*}{Benchmark01 (9 intents)} & \multirow{2}{*}{5} & Flan-T5-XXL & \textbf{86.5} & 0.43 \\
& & GPT-3 & 84.6 & \textbf{0.97} \\
\cline{2-5}
& \multirow{2}{*}{9} & Flan-T5-XXL & 86.5 & 0.48 \\
& & GPT-3 & \textbf{89.3} & \textbf{0.67} \\
\cline{1-5}
\multirow{6}{*}{Benchmark02 (13 intents)} & \multirow{2}{*}{5} & Flan-T5-XXL & \textbf{69.7} & 0.65 \\
& & GPT-3 & 60.6 & \textbf{0.87} \\
\cline{2-5}
& \multirow{2}{*}{13} & Flan-T5-XXL & \textbf{69} & \textbf{0.7} \\
& & GPT-3 & 61.3 & 0.67 \\
\cline{1-5}

\end{tabular}
\caption{Results for zero-shot prediction on 3 internal datasets along with MASSIVE with GPT-3 and Flan-T5-XXL. In-scope accuracy is the accuracy computed for test samples that belong to the intents in the dataset. Out-of-scope recall is the fraction of out-of-scope test samples which were correctly identified as irrelevant by the model i.e., not belonging to any of the intents}
\label{tab:zero_shot}
\end{table*}

\subsection{Prompting Zero-shot Prediction}
\begin{tcolorbox}[colback=gray!10, colframe=black, arc=5pt, label=zero_shot_prompt]
\footnotesize
The given sentence needs to be mapped to exactly one of the intents described below:

\textbf{alarm\_set}: user wants to set an alarm \\
\textbf{iot\_cleaning}: user wants to do some cleaning \\
$\vdots$ \\
\textbf{play\_podcasts}: user wants to play a podcast or rewind/repeat a particular episode in a podcast\\
\textbf{none\_of\_the\_above}: if the user sentence is not about any of the intents above\\
\textbf{Sentence}: wake me up at 7am \\
\sethlcolor{green}
\textbf{Intent}: \hl{\textbf{alarm\_set}}
\label{fig: zero_shot_ex}
\end{tcolorbox}

We use intent names and descriptions for prompting language models to perform zero-shot prediction. The intent descriptions are prefaced with instructions to predict the correct intent and the test utterance is specified at the end. The output is expected to be the correct intent label. Figure \ref{zero_shot_prompt} shows an example prompt from the MASSIVE dataset and the output from LLMs. For evaluation we check for the presence of intent names in the LM completion text as opposed to an exact match and report the first intent predicted in the completion. This is done to account for hallucinations. If no intent names are present in the completion text, we mark it as an ``out-of-scope prediction''. We create intent descriptions for 4 datasets: 3 internal benchmaring datasets and the open-source MASSIVE dataset. We benchmark 2 language models using this type of prompt: GPT-3 (175B parameters) and Flan-T5-XXL (11B parameters), an instruction fine-tuned model.
\paragraph{Filtering Intents for LLMs}
Many conversational agents have a lot of intents, sometimes more than 50. As the number of intents increases, the prompt size increases which incurs higher latency and cost (in case of token-based pricing models). To restrict the length of the prompt, we use sentence similarity to retrieve the top-5 intents and only use those 5 intents in the language model prompt. Using sentence similarity needs a few training examples which makes this a few-shot approach. With just 5 examples per intent, we get more than 0.85 recall for Top-5. To restrict the size of the prompt, we do not pass the training examples to the LMs.  
\begin{figure*}[!t]
    \centering
    \footnotesize
    \begin{subfigure}[b]{0.4\textwidth}
        \includegraphics[width=\textwidth]{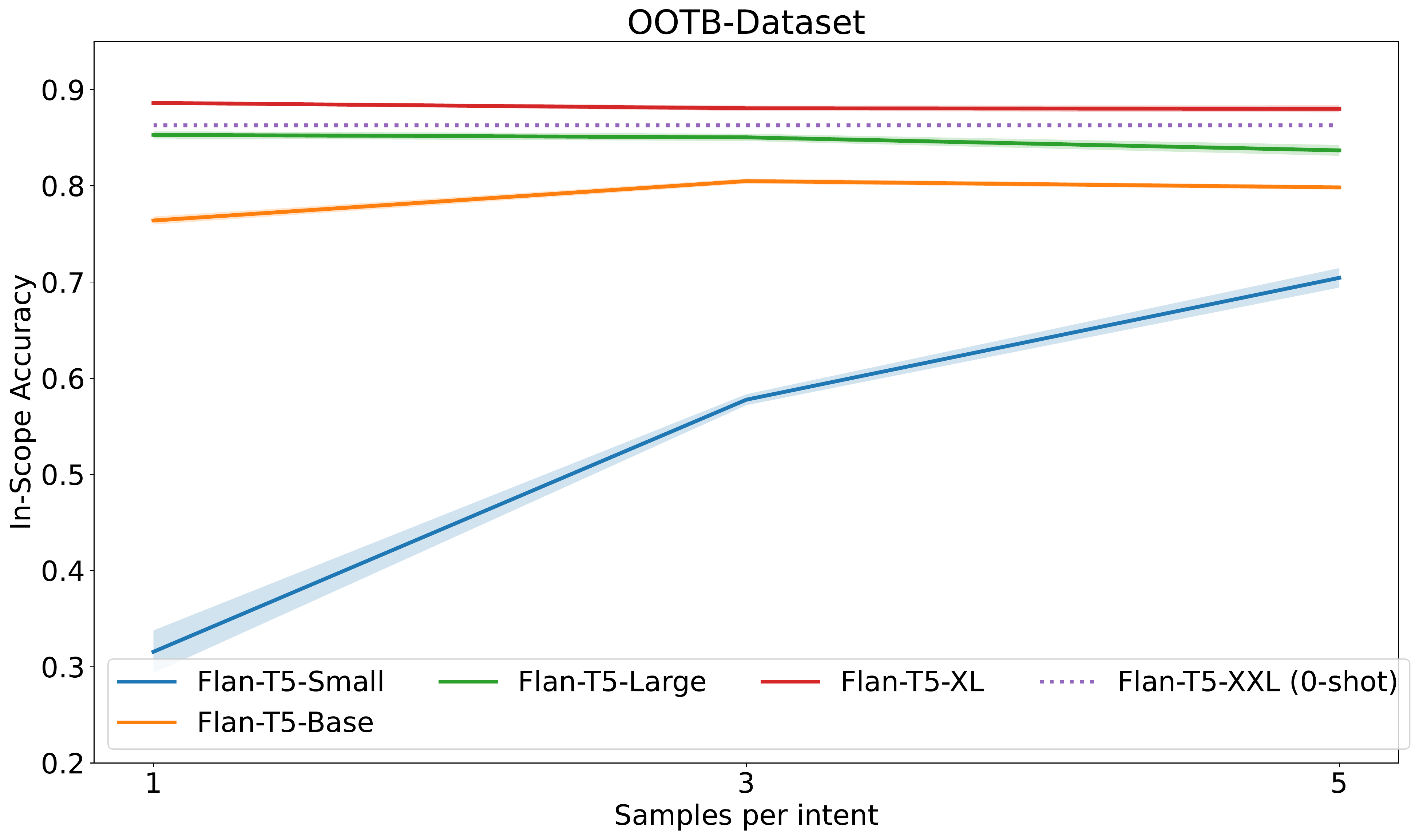}
        \label{fig:plot1}
    \end{subfigure}
    \hspace{0.1\textwidth}
    \begin{subfigure}[b]{0.4\textwidth}
        \includegraphics[width=\textwidth]{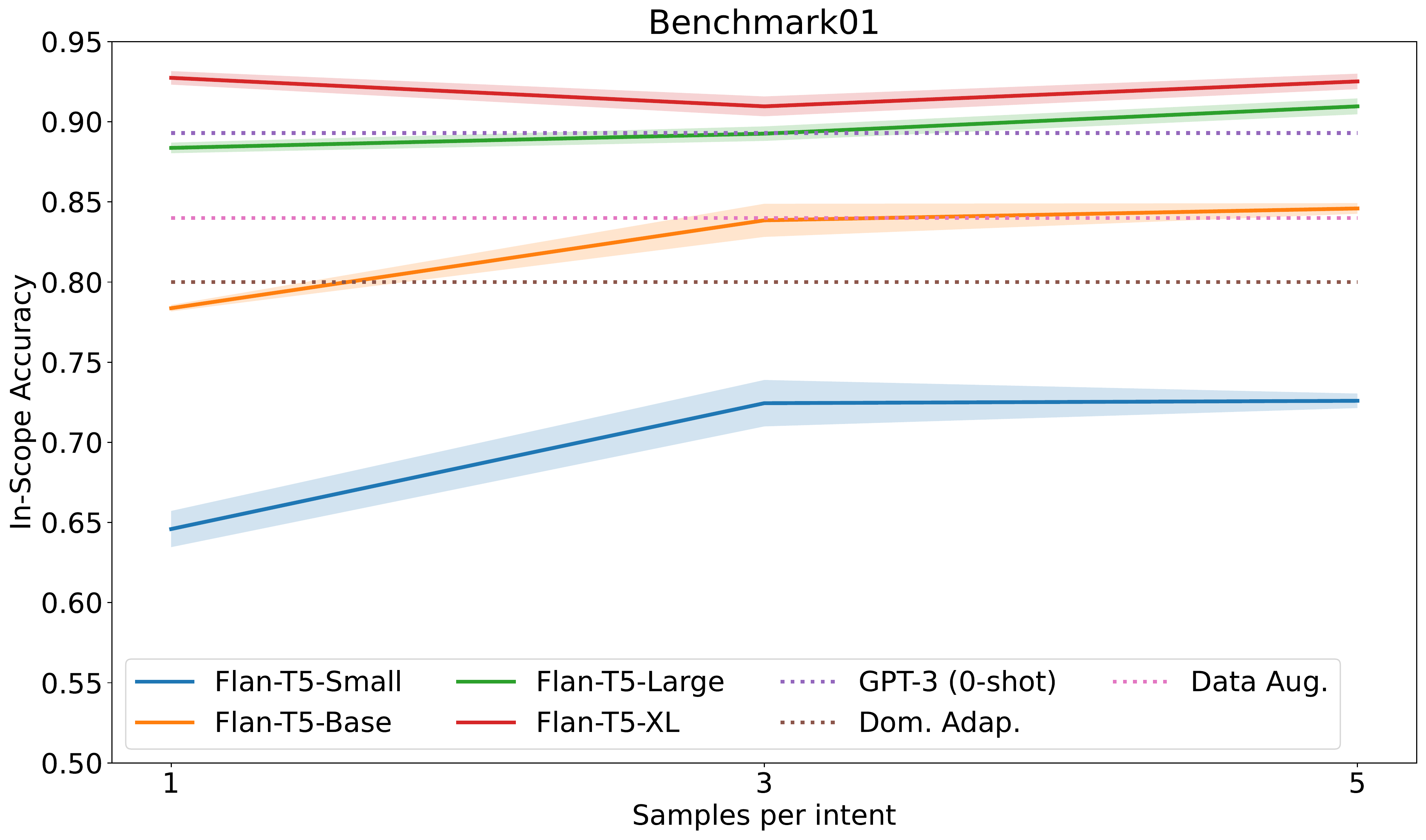}
        \label{fig:plot2}
    \end{subfigure}
    \\
    \begin{subfigure}[b]{0.4\textwidth}
        \includegraphics[width=\textwidth]{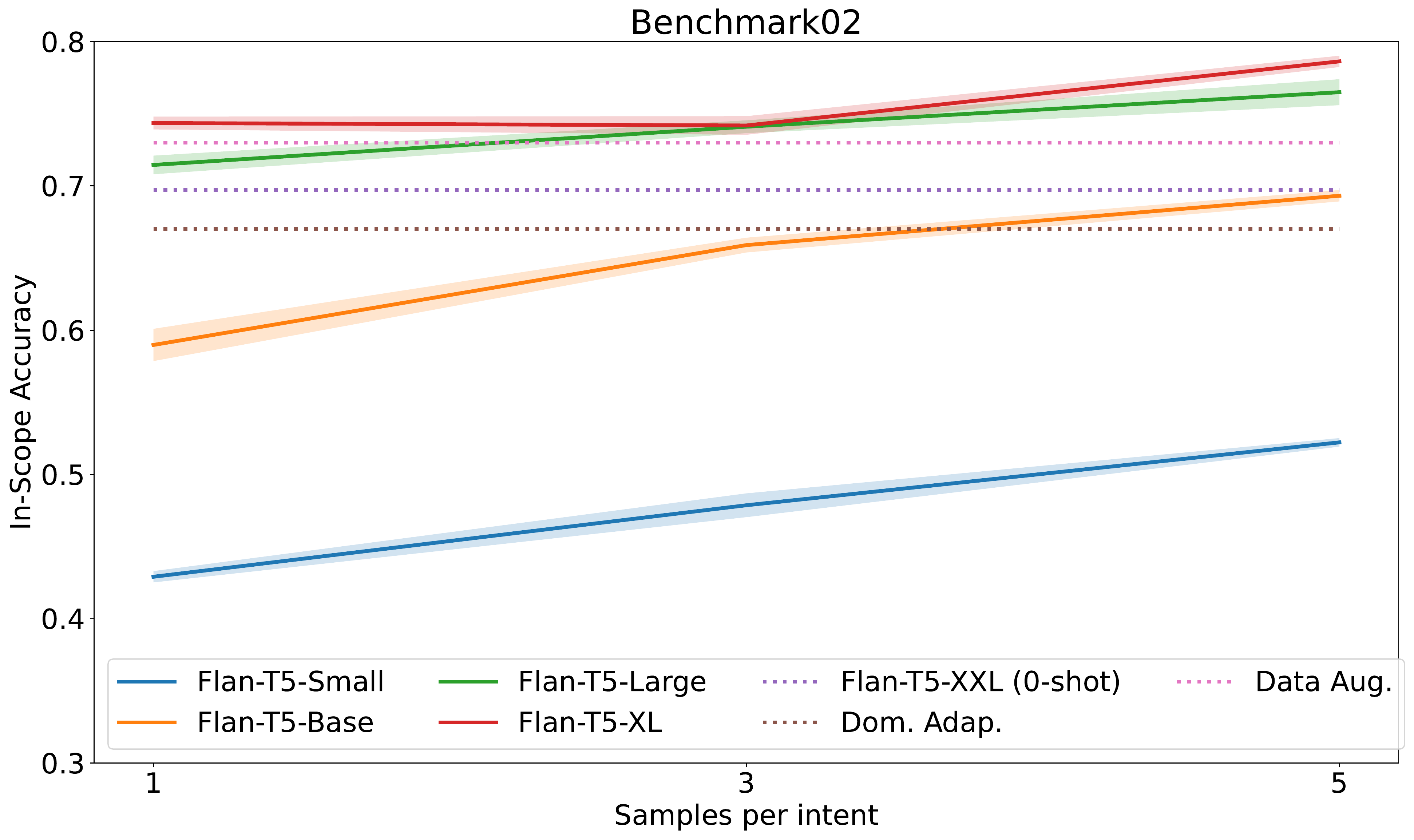}
        \label{fig:plot3}
    \end{subfigure}
    \hspace{0.1\textwidth}
    \begin{subfigure}[b]{0.4\textwidth}
        \includegraphics[width=\textwidth]{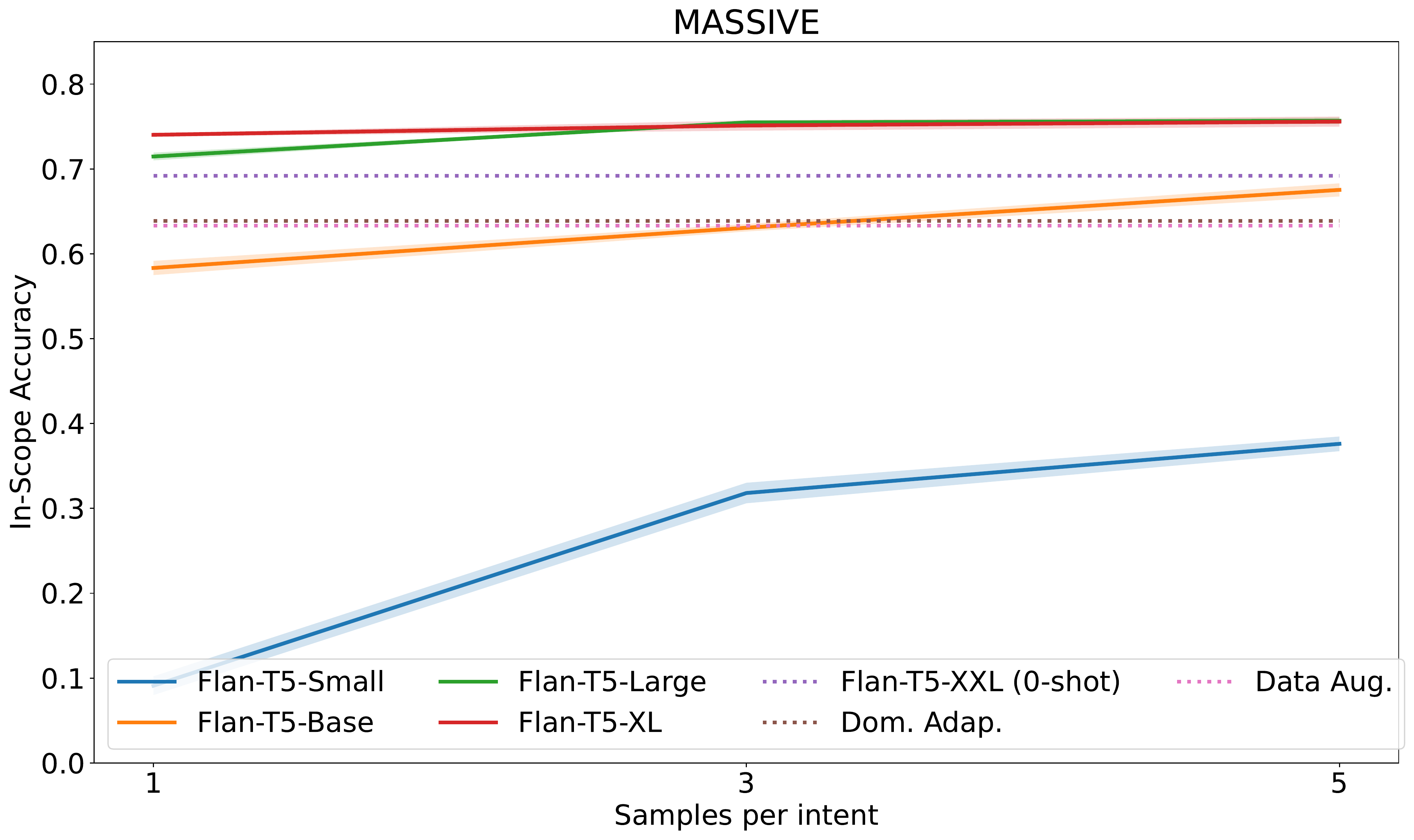}
        \label{fig:plot4}
    \end{subfigure}
    \caption{Plots comparing in-scope accuracy of different Flan-T5 models using Parameter-effiecient FineTuning (PEFT) with the T-Few recipe. The dotted lines show the best results on each dataset from previously described methods. The shaded regions show the standard deviation}
    \label{fig:peft_plots}
\end{figure*}
\paragraph{Setup}
For GPT-3, we set the temperature to 0 and max tokens for generation to 20. We use the default setting generation settings for the Flan-T5-XXL model and do not restrict the number of tokens to be generated. 
The results with filtering are averaged over 3 runs using different random seeds for sampling the 5 samples per intent.  

\paragraph{Results} Table \ref{tab:zero_shot} reports the accuracy for in-scope intents and the recall for out-of-scope samples where applicable (samples that do not belong to any of the intents in the dataset). We find that prompting language models with intent descriptions for zero-shot intent classification performs better than few-shot learning using a classifier (Tables \ref{tab:domain_adaptation_5shot} and \ref{tab:data_augmentation}). Since this only needs intent descriptions, this approach can generalize to new intents as well. Using the same prompt, Flan-T5-XXL is competitive with GPT-3 in terms of in-scope accuracy and is often better when presented a smaller number of intents in the prompt. While the in-scope accuracy is comparable, GPT-3 clearly outperforms Flan-T5-XXL in terms of the out-of-scope recall, indicating that it is better at detecting irrelevant samples. We attribute the strong performance of Flan-T5-XXL (even though it is 16x smaller) to the multi-task instruction finetuning on over 1800 datasets.

For the 3 internal datasets, we also find that using more intents in the prompt works better only up to a certain extent but have excluded the results for the brevity of this paper. While the intent retrieval method does not give perfect Top-5 recall, it helps in keeping the prompt short and hence provides lesser chances for the language models to give a output a wrong label name. Moreover, filtering can also improve the out-of-scope recall as in the case of Benchmark02 dataset.

\subsection{Parameter-Efficient FineTuning (PEFT)}

Taking inspiration from the T-Few recipe \cite{liu2022fewshot}, we add and finetune IA3 adapters from scratch in Flan-T5 models in a few-shot setting which is similar to \ref{subsec:domain_adapt}. We pick K=1,3,5 utterances per intent. Since the Flan-T5 models are instruction fine-tuned, we use the same prompt from \ref{fig: zero_shot_ex} and provide the intent name as the target string. For MASSIVE and OOTB-dataset, we restrict the number of intents in the prompt to 15 at training time to prevent out-of-memory exceptions. At inference time, we provide all intents in the prompt. 
We use all 3 loss functions (language modeling, unlikelihood and length normalized losses) and the same hyperparameters as mentioned in the T-Few paper. For more details about the T-Few recipe, we encourage readers to refer to their paper. 

Figure \ref{fig:peft_plots} compares the results of PEFT against the best results from previously described methods. Flan-T5-XL (3B parameters) consistently outperforms all other methods with just 1 training example per intent. With a few more examples, Flan-T5-Large (770M parameters) also outperforms all other methods except Flan-T5-XXL on the OOTB dataset. This shows that we can train significantly smaller models which are easier to deploy and also outperform LLMs like GPT-3 with just a few parameters using intent descriptions and a handful of examples.
\section{Observations}
\label{sec:observations}
Comparing results across the 4 approaches, we notice that all 4 approaches are effective in low resource settings. We find that domain adaptation is a cheap option in terms of size of the models but it still requires 5-10 training utterances per intent for getting accuracy above 70\%. Data Augmentation using paraphrasing further helps in most cases by 2-4 percentage points. However, expanding to new domains requires sentence-pairs data for training the sentence encoder which can involve days of human labeling.
Zero shot classification using intent descriptions with LLMs and instruction finetuned models performs even better than domain adaptation with data augmentation and doesn't require any utterances to be configured per intent. However a good description for each intent is required. Additionally, these models can be expensive to operationalize. Inference on Flan-T5-XXL requires using A100 GPUs. GPT-3 is not open-source and based on a pricing model which can be expensive to scale to thousands of customers.
Parameter efficient fine-tuning (PEFT) of instruction finetuned models like Flan-T5-XL and Flan-T5-Large offers the best performance across all methods and often by a large margin. Moreover, these models are only a fraction of the size of GPT-3 and Flan-T5-XXL and much easier to operationalize at scale with far lesser compute resources. 

\section{Conclusion}
\label{sec:conclusion}
In this paper, we addressed the task of zero/few-shot intent identification with Large Language Models (LLMs). We presented four approaches, namely domain adaptation, data augmentation, zero-shot prediction with prompting, and parameter-efficient fine-tuning. Our experimental results demonstrate that LLMs and larger instruction fine-tuned language models are very effective in zero-shot setting with in-context prompting. Smaller instruction finetuned models with adapters are even better when adapter-finetuned on just 1 or 3 examples per intent. We hope these results are useful for practical deployment of conversational agents in low-resource settings as well as aiding non-practitioners in building their intent classification models. In the future, we plan to extend this work by domain adapting smaller instruction finetuned models in a multi-task setting and exploring their zero-shot capabiltiies.

\bibliography{main-refs,custom}
\bibliographystyle{acl_natbib}

\end{document}